\documentclass[preprint,1pt,times,twocolumn]{elsarticle}
\usepackage{amssymb}
\usepackage{amsmath}

\usepackage{hyperref}
\usepackage{arydshln}
\usepackage{subcaption}
\usepackage{enumitem}
\usepackage{float}
\usepackage{booktabs}
\usepackage{dblfloatfix}

\newcommand{\fmtC}{\textsc{FMT-C}}
\newcommand{\fmtM}{\textsc{FMT-M}}
\newcommand{\jazzmus}{\textsc{Jazzmus}}
\newcommand{\primus}{\textsc{PrIMuS}}

\newcommand{\assembled}{$\text{CRNN}^{f}$}
\newcommand{\unfolding}{$\text{Unfolding}$}

\newcommand{\smtPrinted}{$\text{SMT}_{\text{pr}}$}
\newcommand{\smtHandwritten}{$\text{SMT}_{\text{hw}}$}

\newcommand\blfootnote[1]{%
  \begingroup
  \renewcommand\thefootnote{}\footnote{#1}%
  \addtocounter{footnote}{-1}%
  \endgroup
}

\journal{Pattern Recognition Letters}

\begin{document}

\twocolumn[{
    \begin{frontmatter}

    \title{Full-Page Optical Music Recognition of Handwritten Monophonic Scores}

    \author[ua]{Adrian Rosello}
    \ead{email address}

    \fntext[fn1]{This is the first author footnote.}
    
    \author[ua]{Antonio Ríos-Vila}
    \author[ua,isea]{David Rizo}
    \author[ua]{Jorge Calvo-Zaragoza}
    
    \affiliation[ua]{organization={Pattern Recognition and Artificial Intelligence group, University of Alicante},
                city={Alicante},
                country={Spain}}
    \affiliation[isea]{organization={Instituto Superior de Enseñanzas Artísticas de la Comunidad Valenciana},
                city={Alicante},
                country={Spain}}
    
    \begin{abstract}
    Full-page end-to-end Optical Music Recognition  seeks to transcribe entire music pages directly into symbolic notation, avoiding the limitations of traditional pipelines that rely on accurate staff segmentation. Recent Transformer-based architectures have achieved strong performance on typeset  scores, relying on large-scale synthetic data for pretraining. However, their applicability to handwritten music remains largely unexplored. In this work, we study full-page transcription on handwritten monophonic collections and analyze the the impact of synthetic pretraining in this setting. To investigate which factors are most relevant during pretraining, we introduce a generator capable of producing visually coherent full-page scores in both typeset and handwritten styles. Experiments on three real handwritten datasets provide a comparative evaluation of several full-page pipelines and different synthetic pretraining strategies. The results suggest that the benefits of synthetic pretraining are primarily associated with learning structural layout conventions rather than with visual similarity to the target handwriting.
    \end{abstract}
    
    
    \end{frontmatter}
}]

\section{Introduction}

Handcrafted sheet music transcription is a tedious process that demands specialized knowledge of the music notation. Yet without the representation obtained from this process, a music document cannot be searched, edited, or analyzed, regardless of whether a digital image of it exists. With recent developments in deep learning, Optical Music Recognition (OMR) has emerged as an alternative to this manual effort, offering a scalable path to digitize the vast collections of music documents that remain inaccessible~\cite{Calvo-Zaragoza:ACM:2020}.

The end goal of OMR is the recognition of full music pages, which requires models to handle both layout understanding and symbol recognition across the entire page. For a long time, this complexity made holistic full-page transcription a challenging task. Music sheets are structured documents composed of multiple staves, each containing a sequence of musical symbols that must be read in order. Recognizing the full page means not only identifying those symbols correctly but also understanding how the staves are arranged and how they relate to each other. Given this difficulty, early approaches converged toward a simpler proxy: a two-step process where layout analysis first locates all staves, and each one is then transcribed individually. This setup, however, creates a snowball effect where layout analysis errors propagate through the system. Furthermore, treating each staff in isolation prevents the model from leveraging inter-staff context, such as consistent key and time signatures. Despite these limitations, staff-level approaches have shown strong recognition performance, with most systems adopting the Convolutional Recurrent Neural Network (CRNN) architecture trained with Connectionist Temporal Classification (CTC) loss~\cite{Calvo-Zaragoza:ISMIR:2018,Villareal:ICFHR:2020,Martinez-Sevilla:ISMIR:2023}, combining convolutional layers for feature extraction with recurrent layers for temporal modeling.

Recent works have taken a step further by exploring holistic full-page approaches that transcribe entire sheets in a single step, removing this dependency altogether. Early efforts in this direction led to the so-called \textit{Unfolding} method~\cite{Martinez-Sevilla:ICDAR:2023}, which trains models to flatten multi-staff pages into sequential representations. However, this flattening causes a loss of spatial context, limiting the method to specific domains. To address this, subsequent works adopted Transformer-based architectures capable of transcribing without spatial constraints. A noteworthy contribution was the work of \citet{rios2025}, who proposed the Sheet Music Transformer (SMT), an encoder-decoder architecture trained with curriculum learning on large-scale synthetic data, achieving state-of-the-art full-page transcription even under limited data conditions.

Yet, these advances have been largely confined to very specific scenarios. The Unfolding method has been primarily evaluated on mensural notation, a historically simpler form of music that does not reflect the notational complexity of modern scores. The SMT, on the other hand, has been designed and optimized for polyphonic typeset modern music, leaving other sources and styles largely unexplored. This raises a  question: how well do these methods generalize beyond the narrow settings in which they were developed?

\begin{figure}[b]
    \centering
    \includegraphics[width=.98\linewidth]{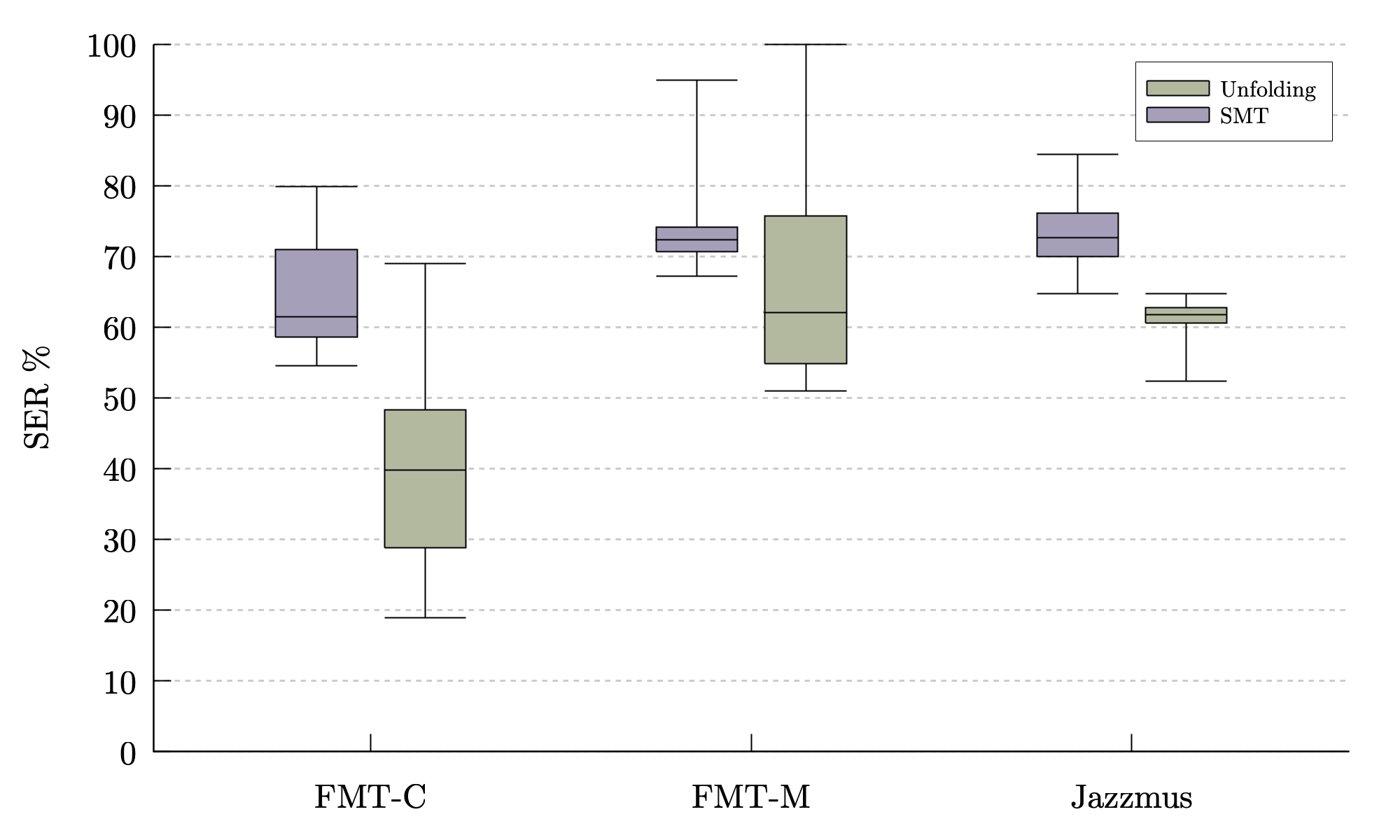}
    \caption{10-fold error rates after directly adjusting the \unfolding~\cite{Martinez-Sevilla:ICDAR:2023} and SMT~\cite{rios2025} models to the target handwritten corpora.}
    \label{fig:robustness}
\end{figure}

Indeed, this concern is backed by our preliminary experiments. When applying these methods directly to handwritten monophonic sources, both the Unfolding and SMT models fail to transfer successfully, even after fine-tuning on the target domain (Fig.~\ref{fig:robustness}). This gap demonstrates that full-page models cannot be directly applied to new scenarios without careful consideration of the target domain.

Although the effect of handwritten data on full-page models remains largely unknown, thanks to recent advances in synthetic music generation, it is now possible to study musical styles and textures that have never been explored in the context of full-page recognition. For instance, the Smashcima~\cite{smashcima} generator, which has not yet been studied in an experimental setting, now enables the generation of realistic handwritten music scores, opening new frontiers to study this phenomenon at a larger scale.

In this work, we present the first study of full-page OMR on handwritten monophonic collections. We investigate whether the difficulty of this transfer stems from the visual variability of handwritten notation or from the structural properties of monophonic music. To this end, we introduce a tailored full-page synthetic generator capable of producing scores in both typeset and handwritten styles, enabling controlled pretraining experiments that reveal what truly matters when adapting full-page models to this unexplored scenario.\blfootnote{The code of this paper will be released upon acceptance of this paper. This footnote will be changed to reflect the link to the online code repository in GitHub.}

\section{Methodology} \label{sec:methodology}

The goal of full-page transcription is to, given a set of full-page images $\mathcal{X}$, retrieve the sequence of tokens $\mathbf{t} = \left(t_{1},t_{2},\ldots,t_{n}\right)$ of the music embedded in a single image $x\in\mathcal{X}$. Each possible value of any token $t_i$ is represented in a finite vocabulary $\Sigma$, being $\Sigma^*$ all possible combinations of sequences from this vocabulary.

We assume that there exists a function $f : \mathcal{X} \rightarrow \Sigma^{*}$ that relates each sheet image $x\in\mathcal{X}$ with its sequence representation $t\in\Sigma^*$. Evaluated techniques will try to approximate this function $f$ as $\hat{f}$ by selecting the most probable token at each position in the sequence given any input image $x \in \mathcal{X}$. The way this likelihood is computed depends on the decoding strategy:

\begin{itemize}[nosep, leftmargin=10pt, label={}]
    \item \textbf{CTC greedy decoding.}
    In CTC-based models, the output of the network is a sequence of frame-wise distributions over an extended vocabulary $\Sigma'\!=\!\Sigma\,\cup\,\varnothing$, which includes a blank symbol. Given an input image $x$, the model produces a sequence of predictions $\pi~=~(\pi_1, \pi_2\ldots, \pi_k)$ of length $K$, where each $\pi_k \in \Sigma'$ is chosen independently of the vocabulary $\Sigma'$ based on its predicted likelihood at frame $k$. The final transcription $t$ is obtained by applying the CTC collapsing function $\mathcal{B}\left(\cdot\right)$:
    \begin{equation*}
    \hat{f}(x) = \mathcal{B}(\pi), \text{with} ~ \pi_k = \arg\max_{\pi \in \Sigma^*}  P(\pi \mid x, k)
    \end{equation*}
    Here, $k$ indexes the frame-wise outputs of the network, while the resulting sequence $t$ of length $n \leq K$ contains the actual predicted tokens after merging repeated symbols and removing blanks.

    \vspace{1em}

    \item \textbf{Autoregressive decoding.}
    In autoregressive models, the sequence is generated token by token, each conditioned on the input image and all previously predicted tokens. The model calculates the output probability as a product of conditional probabilities:
    \begin{equation*}
    \hat{f}(x) = (\hat{t}_1, \ldots,\hat{t}_n), \text{with} ~ \hat{t}_i = \arg\max_{t \in \Sigma^*} P(t \mid x, \hat{t}_{<\,i})
    \end{equation*}
    
    \noindent where $\hat{t}_{<\,i} = (\hat{t}_1, \ldots, \hat{t}_{i-1})$ is the predicted prefix up to position $i-1$. During training, the model is optimized using a teacher forcing policy, where the true prefix $t_{<\,i}$ is used instead.
\end{itemize}

\section{Experimental Setup} \label{sec:setup}

This section describes the experimental setup, including the evaluated pipelines, the synthetic data generation process, the datasets, and the evaluation protocol.

\subsection{Two-steps Pipeline} \label{sec:staff-line}

To compare with staff-level architectures, we adopt a segmentation-based strategy, which we will refer to from now on as \assembled. This approach applies first a layout analysis with YOLOv11~\cite{Dvorak2024} to segment the input page into individual staves, which are then concatenated into a single sequence for the CRNN model. However, this concatenation produces significantly longer sequences than those in standard staff-level training. Since the CRNN uses a bidirectional LSTM with limited long-range dependency modeling, this setup is suboptimal. Thus, \assembled~provides a fair but constrained adaptation of CRNN to full-page recognition.

\begin{figure}[h]
    \centering
    \includegraphics[width=.85\linewidth]{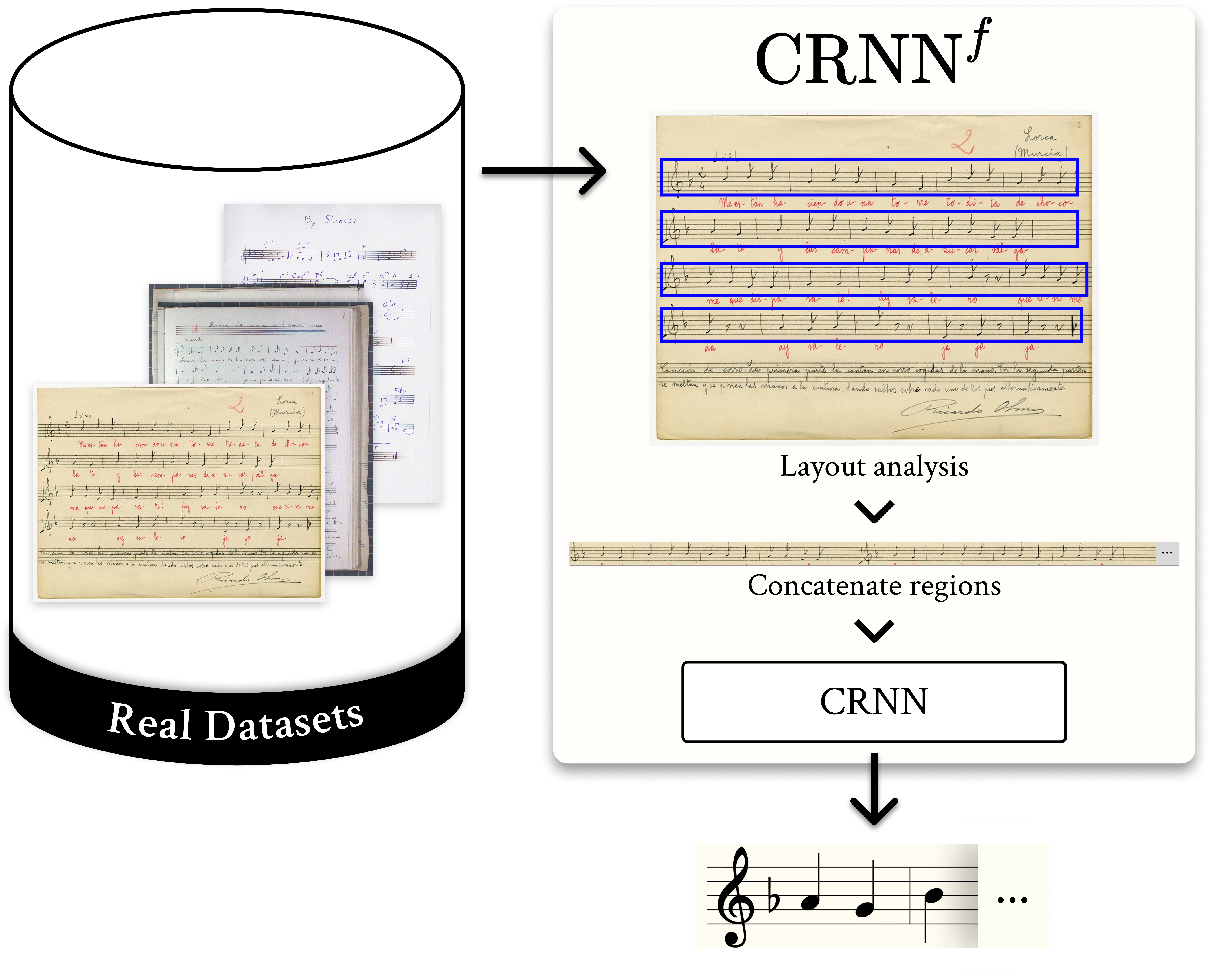}
    \caption{Diagram of the two-step pipeline used.}
    \label{fig:two-steps-pipeline}
\end{figure}

\subsection{Holistic Pipelines}

We evaluate two segmentation-free full-page approaches: the \unfolding~method~\cite{Martinez-Sevilla:ICDAR:2023}, trained directly on the target corpora, and the SMT architecture~\cite{rios2025}, which follows a curriculum-based pretraining strategy on synthetic data of incremental difficulty before fine-tuning on the real datasets.

\begin{figure}[h]
    \centering
    \includegraphics[width=\linewidth]{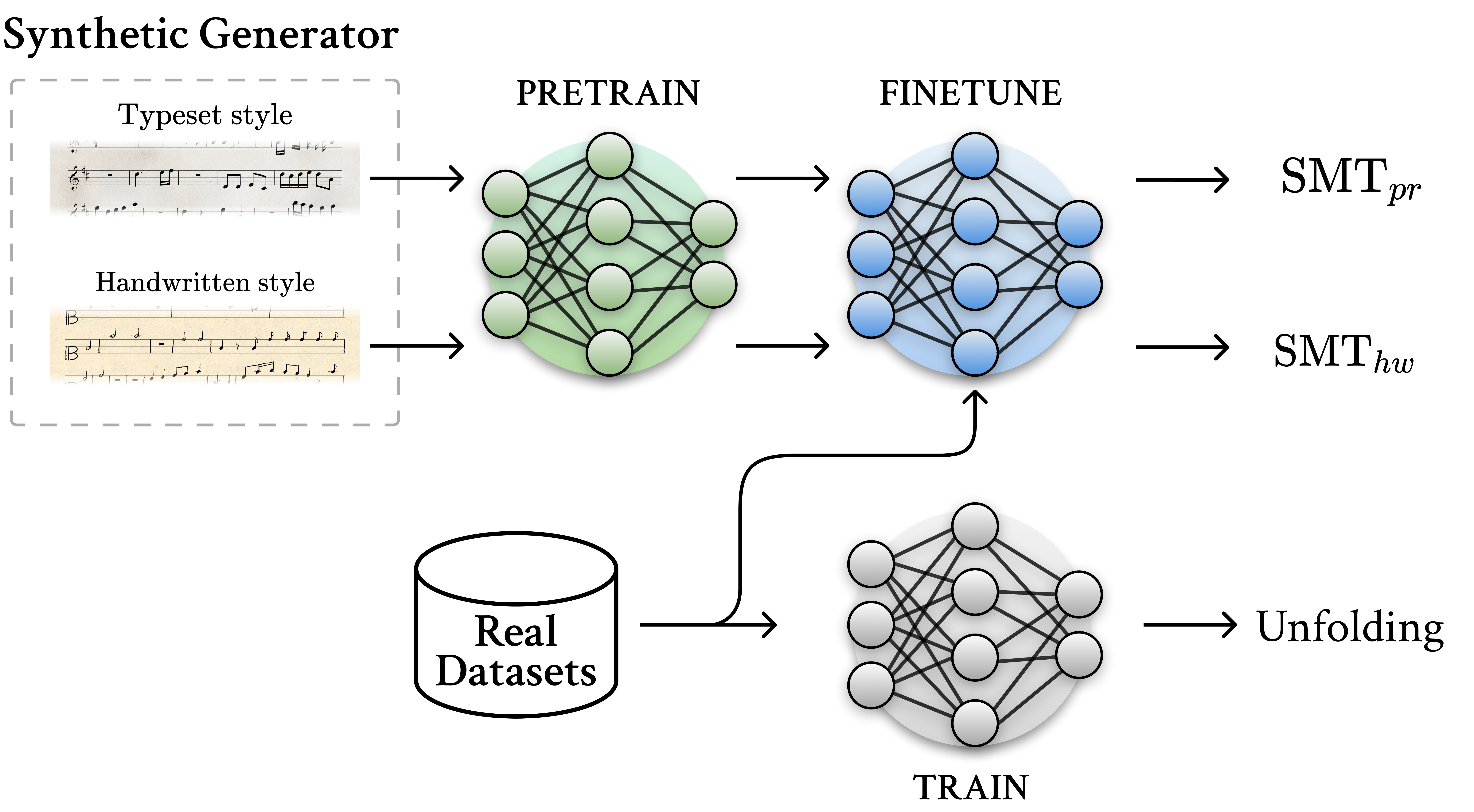}
    \caption{Diagrams of the holistic pipelines used.}
    \label{fig:holistic-pipeline}
\end{figure}

\subsection{Label Representation}

The choice of label representation is a trade-off between vocabulary complexity and sequence length. While agnostic encodings focus on graphical primitives, semantic representations like MusicXML or MEI \cite{Good2001MusicXMLAI, Hankinson2011TheME} capture musical meaning but are often too verbose for sequence-to-sequence modeling.

In this work, we adopt the basic extended \texttt{**kern} (\texttt{**bekern}) representation~\cite{rios2025}, which decomposes \texttt{**kern} tokens into atomic components separated by delimiters. Conversion from \texttt{**bekern} to \texttt{**kern} is fully reversible through naive preprocessing tools~\cite{kernpy_mec_2025}, ensuring compatibility with existing resources.

\subsection{Synthetic Data Generation}

To pretrain the SMT architecture, we introduce a full-page generator tailored for synthetic monophonic sheets, built on the RISM incipits that serve as the source for the \primus~corpus~\cite{Calvo-Zaragoza:ISMIR:2018}.

\begin{figure}[h]
    \begin{subfigure}[b]{0.45\linewidth}
        \centering
        \includegraphics[width=\linewidth]{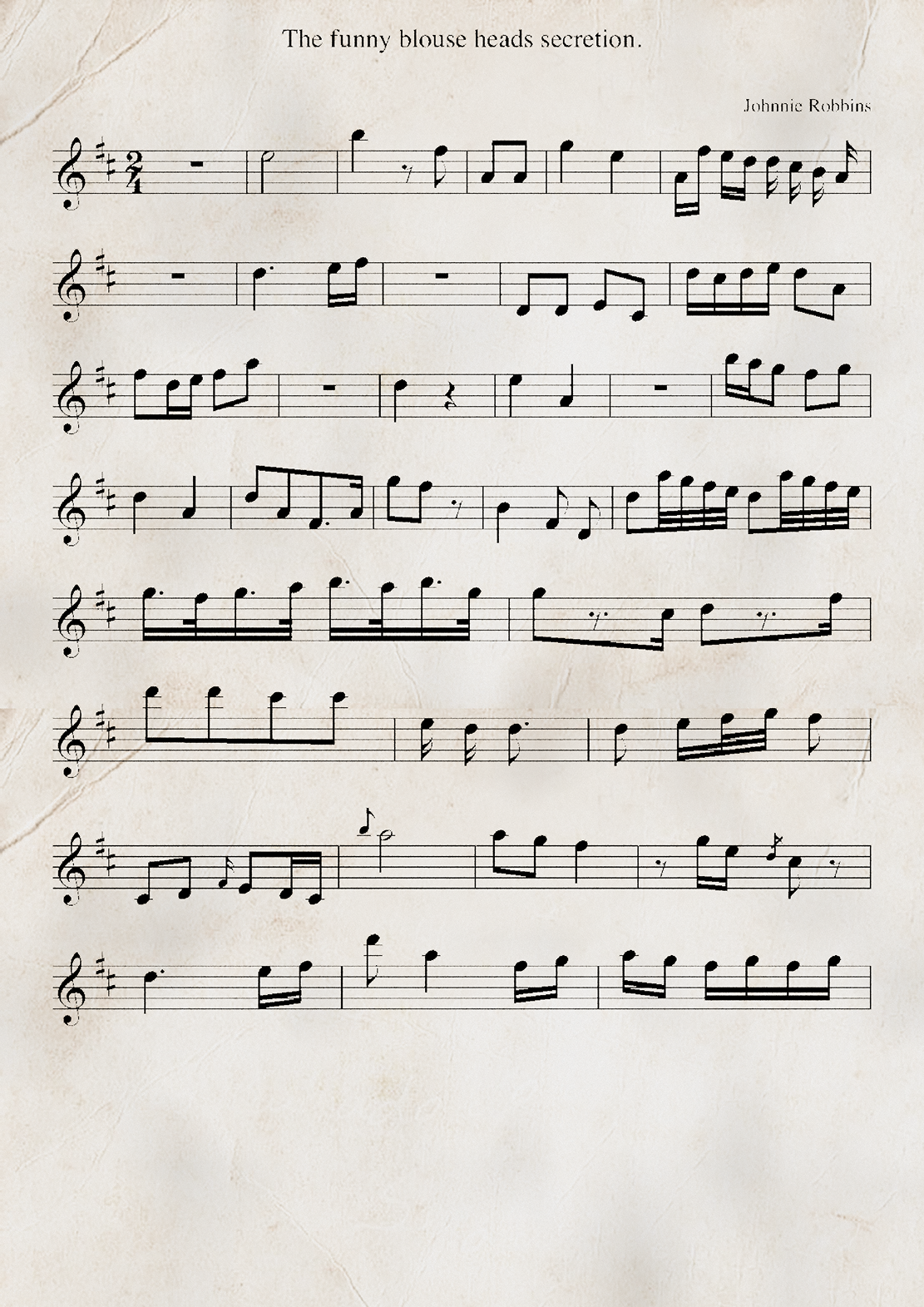}
        \caption{Typeset style}
        \label{fig:pr-style}
    \end{subfigure}
    \hfill
    \begin{subfigure}[b]{0.45\linewidth}
        \centering
        \includegraphics[width=\linewidth]{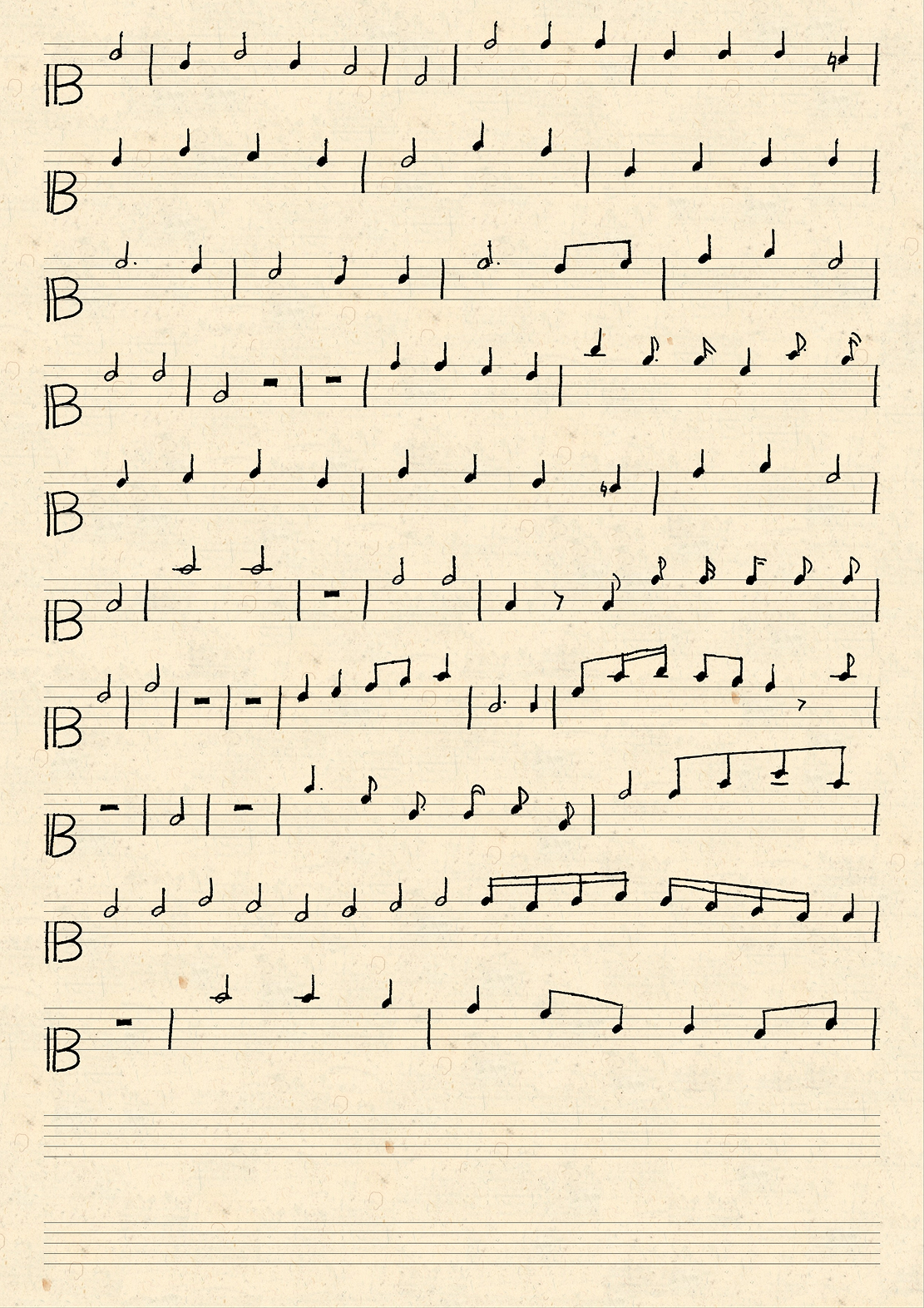}
        \caption{Handwritten style}
        \label{fig:hw-style}
    \end{subfigure}
    \caption{Representative examples of the generated music in both typeset (a) and handwritten (b) style.}
    \label{fig:synth-styles}
\end{figure}

Combining RISM staves naively into full pages can produce notational inconsistencies across staves---such as mismatched clefs, key signatures, or time signatures---leading to visual artifacts in the rendered output. To address this, the generator segments each incipit by measure and indexes them according to three musical attributes: key signature, time signature, and clef. These attributes are enforced to remain constant across all staves on a page, preventing artifacts such as notes rendered outside the staff range. Once a set of measures has been selected, a synthetic page is generated in either typeset or handwritten style (Fig.\ref{fig:synth-styles}). The goal is to produce musically coherent scores that are consistent from a visual and syntactic conventions point of view, not to generate musically meaningful compositions in any specific genre or style.

\subsubsection{Typeset Style} \label{sec:typ}

To simulate a typeset style, we use Verovio~\cite{Pugin:ismir:2014} engraver. The toolkit provided by the engraver enables various modifications to the layout to generate different compositions. We can leverage this to provide randomization of such layouts to create more realistic distributions. To this end, applied variations modify page margins, content scale, and inter-system spacing. The engraver also allows creating textual elements like the work title or the author name, which we can also randomize using available text generators.

By default, the pages yielded by this engraver are retrieved in plain black and white. To increase the resemblance to real manuscripts, a background texture is selected from a subset of images and then added to the scores. Even with the background textures, images are still far from realistic. For this reason, a set of transformations is applied using Augraphy~\cite{augraphy_paper}. Augraphy is a library that simulates paper-handling distortions by generating effects like ink bleeding, letterpress, ink shift, and background texturization. After this process, the generated pages achieve a visual quality much closer to scanned historical prints.

\subsubsection{Handwritten Style} \label{sec:hw}

For obtaining a handwritten style, we rely on the Smashcima engraver~\cite{smashcima}, a synthesizer specifically designed to generate full-page manuscript-style scores from symbolic encodings. Unlike traditional engravers, Smashcima produces glyphs that imitate the irregularities and spatial variability characteristic of handwritten notation. Given a MusicXML input, it renders a complete page while preserving detailed semantic information, including glyph locations, staff associations, and notational structure. Since the original RISM data is encoded in **kern, we convert the synthesized output into MusicXML using the converter21\footnote{\href{https://github.com/gregchapman-dev/converter21}{https://github.com/gregchapman-dev/converter21}} python library, a tool built on music21\footnote{\href{https://www.music21.org/music21docs/}{https://www.music21.org/music21docs/}} that provides bidirectional converters between many semantic encodings. We use the converted MusicXML to generate the synthetic image and leverage the **kern sequence to obtain the label sequence.

\subsection{Evaluation Corpora}

To assess the performance of the proposed strategies under realistic conditions, we evaluate them on three real full-page music datasets~\footnote{Due to the resolution of full-page images, examples of each dataset are depicted in the supplementary material.}. These collections differ in layout, writing style, and level of degradation, offering a representative benchmark.

\begin{itemize}[nosep,leftmargin=5pt, label={-}]
    \item \jazzmus~corpus~\cite{martinezsevilla2025opticalmusicrecognitionjazz}: a collection of 293 handwritten jazz lead sheets gathered from students and professional musicians, characterized by irregular handwriting styles, strike-through corrections, diverse capture conditions, and a wide variety of chord symbol conventions.

    \item \fmtC~and \fmtM~corpora~\cite{fmt}: two partitions derived from a collection of four groups of handwritten score sheets of popular Spanish songs, notated in modern music notation. The source material originates from the \emph{Fondo de Música Tradicional} (IMF-CSIC). \fmtC~comprises vertically oriented images with a significant presence of textual elements, while \fmtM~consists of landscape-format pages that exhibit signs of paper aging, such as stains and yellowing.
\end{itemize} 

\begin{table}[ht]
    \centering
    \setlength\tabcolsep{2em}
    \resizebox{\linewidth}{!}{
        \begin{tabular}{lccc}
            \toprule[1pt]
            \textbf{Name} & \fmtC & \fmtM & \jazzmus \\
            \midrule
             \textbf{Vocabulary size} & 60 & 75 & 96 \\
             \textbf{Number of pages} & 86 & 205 & 163 \\
            \bottomrule[1pt]
        \end{tabular}
    }
    \caption{Statistics of the real full-page datasets.}
    \label{tab:dataset_stats}
\end{table}

\subsection{Evaluation Protocol}

For evaluating the performance of the benchmarked models in full-page transcription, we resorted to the Symbol Error Rate (SER) figure of merit. This metric, widely used in previous works~\cite{Baro:ICFHR:2020,Villareal:ICFHR:2020,Alfaro-Contreras:PRL:2022}, measures the minimum number of operations---referring as operations to insertions, modifications, or eliminations of symbols---required to match the sequence predicted by the models to that of the ground truth.

Due to the limited size of the real full-page datasets (Table~\ref{tab:dataset_stats}), we adopt a 10-fold cross-validation strategy to ensure a fair comparison. Results are reported as the average SER across all folds, along with the standard deviation to capture variability.

\begin{figure*}[!b]
    \centering
    \includegraphics[width=\linewidth]{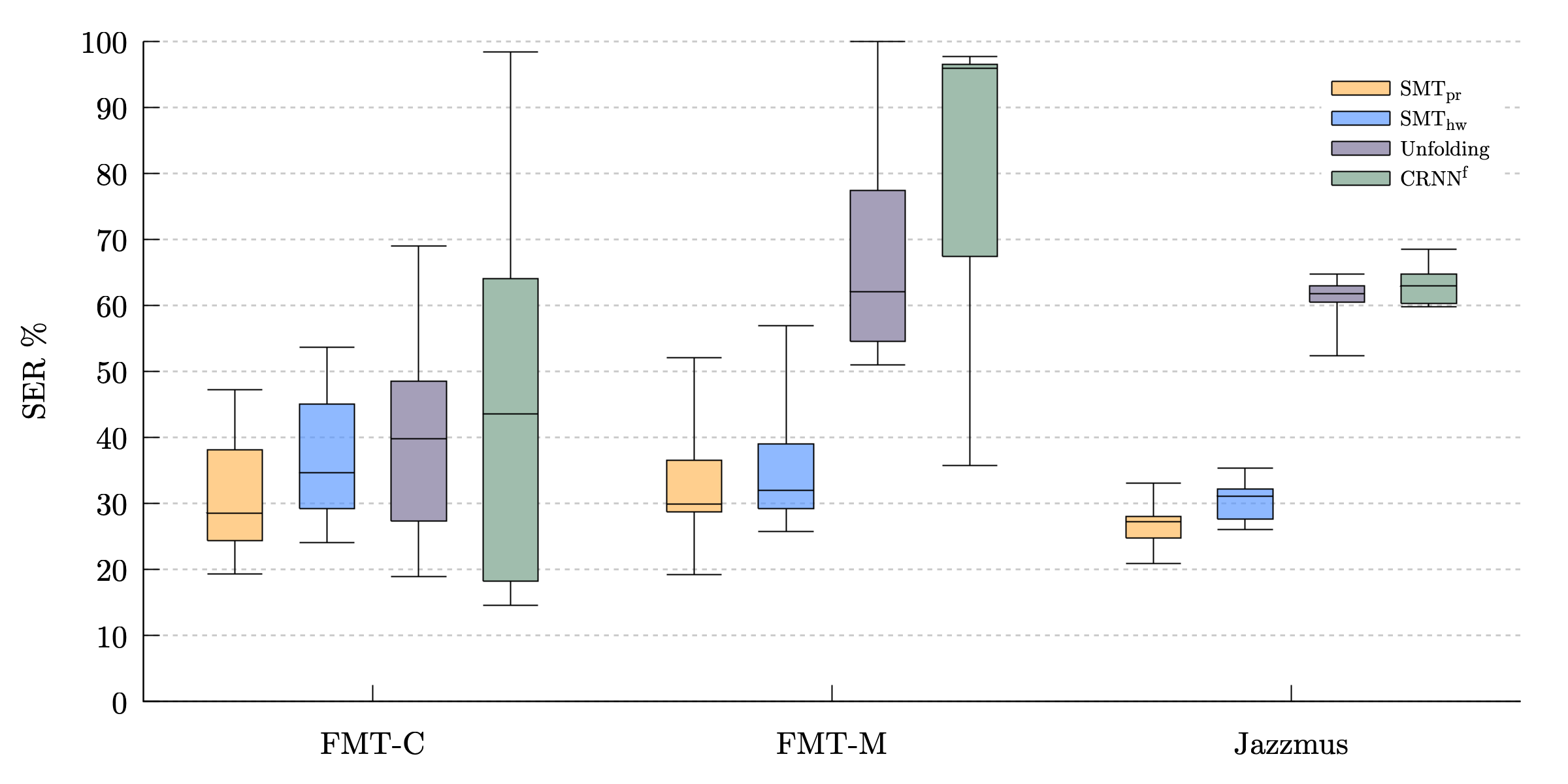}
    \caption{Comparison of the SMT, Unfolding, and \assembled scenarios in terms of SER across 10 folds for the \fmtC, \fmtM, and \jazzmus~datasets.}
    \label{fig:whiskers}
\end{figure*}

\section{Results} \label{sec:results}

The experimental outcomes presented in Table~\ref{tab:avg_results}, indicate that the SMT architecture consistently outperforms other baselines across all tested corpora regardless of the pretraining data style.  

\begin{table}[h]
    \centering
    \setlength\tabcolsep{1.5em}
    \resizebox{\linewidth}{!}{
    \begin{tabular}{lrccc}
        \toprule[1pt]
                      && \fmtC                           & \fmtM                        & \jazzmus                  \\ \midrule
    & \assembled       & $45.7\pm27.8$                   & $81.6\pm25.4$                & $63.1\pm2.8$              \\
    & \unfolding       & $40.4\pm14.9$                   & $68.3\pm18.0$                & $61.0\pm3.5$              \\
    & \smtPrinted      & $\mathbf{31.6}\pm\mathbf{9.9}$  & $\mathbf{32.8}\pm8.7$        & $\mathbf{26.9}\pm3.4$     \\
    & \smtHandwritten  & $38.4\pm10.4$                   & $32.9\pm\mathbf{5.2}$        & $30.6\pm\mathbf{3.1}$     \\
    \cmidrule{1-5}
    \end{tabular}
    }
    \caption{10-fold cross-validated SER (\%) on \fmtC, \fmtM, and \jazzmus. Values report mean and standard deviation across folds. Best results per dataset are in bold.}
    \label{tab:avg_results}
\end{table}

A more in-depth plot of the 10-fold evaluation is provided in Figure~\ref{fig:whiskers}, where we can observe that, while the SMT models maintain a low average error, the \assembled~and \unfolding~systems exhibit highly unstable behavior. In several folds, these baseline systems reach extreme error rates, whereas the SMT models generalize more reliably to varying layout conditions. However, the \unfolding~approach remains competitive in specific monophonic contexts, as we can see by the results obtained in~\fmtC, comparable to those obtained by~\smtHandwritten.

Regarding the influence of synthetic pretraining, the research reveals a disparity between visual fidelity and model accuracy. Although \smtHandwritten~was developed to resemble closelly the handwritten domain, \smtPrinted~yielded a lower SER in both the \fmtC~and \jazzmus~collections, while achieving a marginal performance difference in the \fmtM~dataset. These findings indicate that the primary contribution of synthetic pretraining lies in the acquisition of structural and semantic priors rather than the imitation of specific handwritten styles.

\subsection{Qualitative Evaluation}

\begin{figure}[!h]
    \begin{subfigure}[b]{0.47\linewidth}
        \centering
        \includegraphics[width=\linewidth]{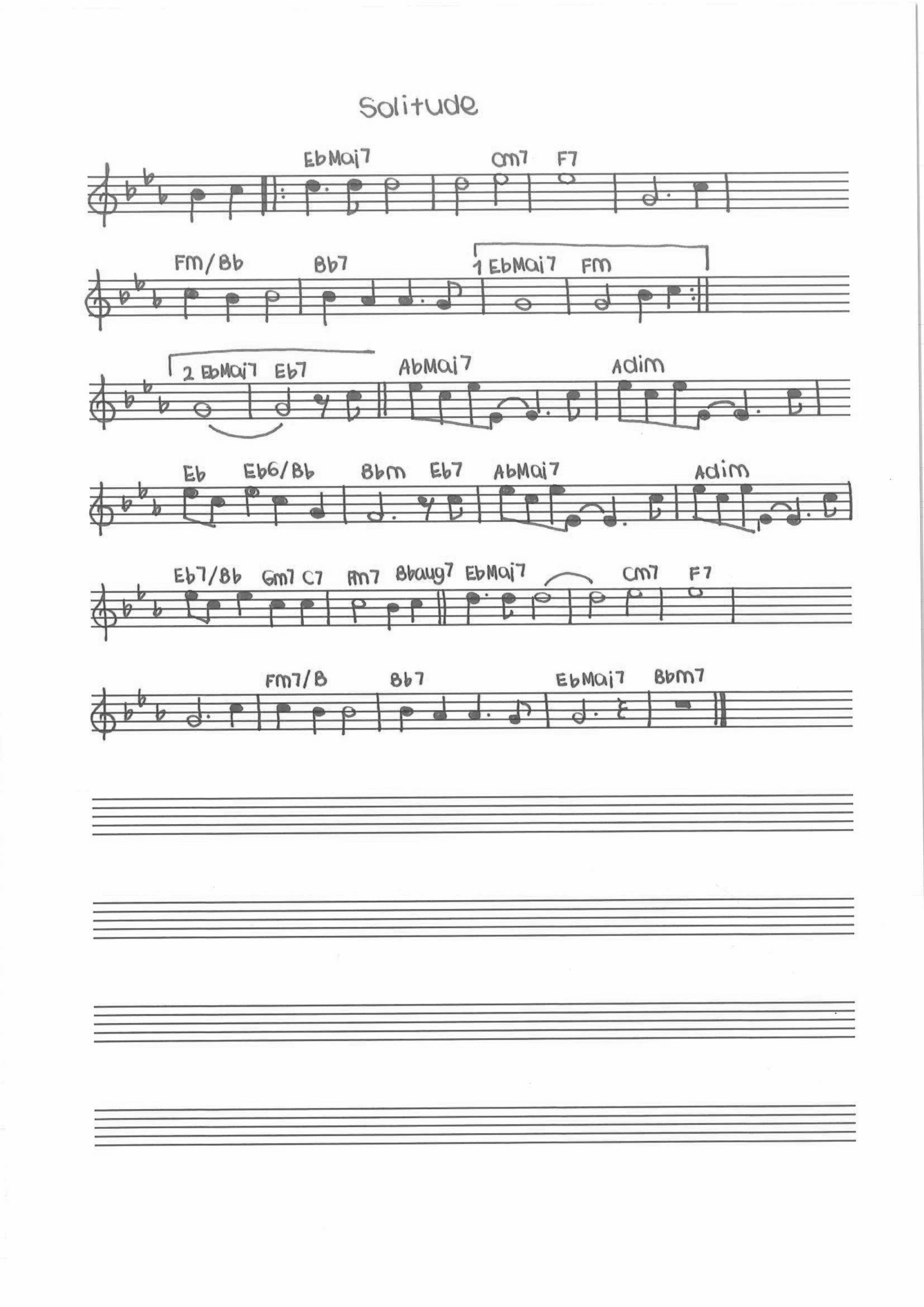}
    \end{subfigure}
    \hfill
    \begin{subfigure}[b]{0.47\linewidth}
        \centering
        \includegraphics[width=\linewidth]{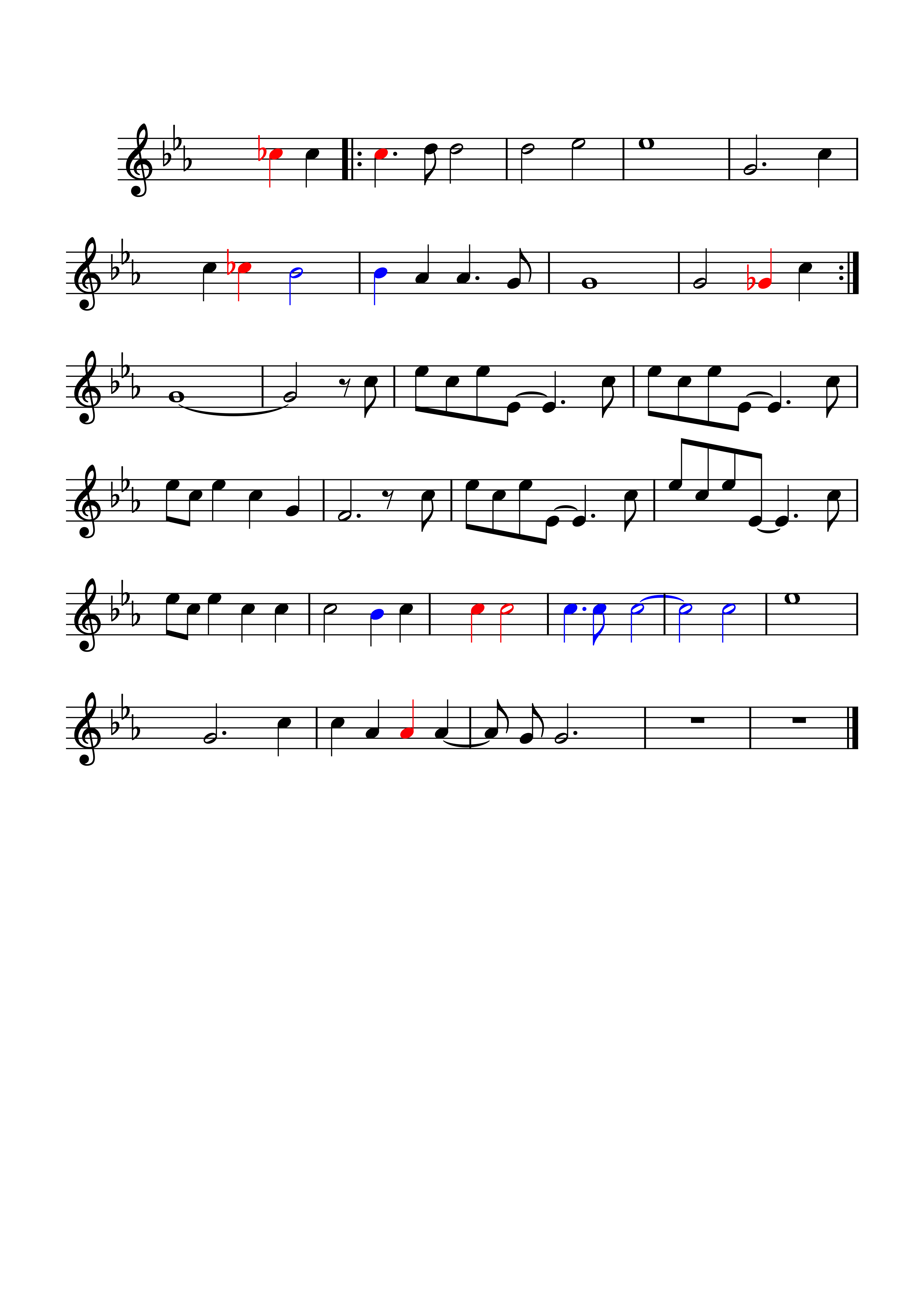}
    \end{subfigure}
    \caption{Qualitative result of the model \smtPrinted~performance on a test split of the 10k-fold of \jazzmus. Red denotes false positives, while blue represents false negatives.}
    \label{fig:qualitative}
\end{figure}

Figure~\ref{fig:qualitative} shows the output of \smtPrinted\ on a sample from \jazzmus. The model correctly discards irrelevant content such as empty staves and the title heading, and recovers after local failures, resuming accurate transcription in subsequent measures. Most false negatives arise from non-renderable syntactic errors; false positives correspond to pitch misassignments.

Locating these errors, however, is not straightforward. Mistakes are scattered across long symbolic sequences spanning multiple staves, making it difficult to assess where and why the model fails. Despite this being a central challenge in the digitization of music scores, no alternative to careful manual review currently exists.

\section{Conclusions} \label{sec:conclusions}

This research addressed the challenge of graphical variability in handwritten musical heritage. The SMT architecture consistently outperformed both the segmentation-based and unfolding baselines across three datasets with different layout conditions, as confirmed by 10-fold cross-validation. The other pipelines not only produced higher error rates but also showed large variation across folds.

A key finding is that typeset pretraining matched or surpassed handwritten pretraining on all datasets, meaning that what matters most in synthetic data is page layout and notation conventions, not visual resemblance to handwriting. Future work should build on this premise when considering similar pretraining strategies.

\section{Acknowledgements}
Work supported by grant CISEJI/2023/9 from ``Programa para el apoyo a personas investigadoras con talento (Plan GenT) de la Generalitat Valenciana''. 

The first author is supported by grant CIACIF/2024/372 from ``Programa I+D+i de la Generalitat Valenciana''.

\bibliography{bibliography}

\end{document}